\documentclass{article}

\usepackage{microtype}
\usepackage{graphicx}
\usepackage{subfigure}
\usepackage{booktabs} 

\usepackage{hyperref}

\usepackage[accepted]{icml2025}

\usepackage{amsmath}
\usepackage{amssymb}
\usepackage{mathtools}
\usepackage{amsthm}
\usepackage{algpseudocode}
\usepackage{algorithm}
\usepackage{fancyhdr}
\usepackage[capitalize,noabbrev]{cleveref}

\theoremstyle{plain}

\theoremstyle{definition}

\theoremstyle{remark}

\usepackage[textsize=tiny]{todonotes}

\icmltitlerunning{Enhancing Math Reasoning in Small-sized LLMs}

\begin{document}

\twocolumn[
\icmltitle{Enhancing Math Reasoning in Small-sized LLMs via Preview Difficulty-Aware Intervention}

\icmlsetsymbol{equal}{*}

\begin{icmlauthorlist}
\icmlauthor{Xinhan Di}{yyy}
\icmlauthor{JoyJiaoW}{comp}

\end{icmlauthorlist}

\icmlaffiliation{yyy}{Independent, Researcher, China}
\icmlaffiliation{comp}{DeepReason, Shanghai, China}

\icmlcorrespondingauthor{JoyJiaoW}{deepreasoninggo@gmail.com}

\icmlkeywords{Machine Learning, ICML}

\vskip 0.3in
]

\printAffiliationsAndNotice{\icmlEqualContribution}
\begin{abstract}
Reinforcement learning scaling enhances the reasoning capabilities of large language models, with reinforcement learning serving as the key technique to draw out complex reasoning. However, key technical details of state-of-the-art reasoning LLMs—such as those in the OpenAI O series, Claude 3 series, DeepMind’s Gemini 2.5 series, and Grok 3 series—remain undisclosed, making it difficult for the research community to replicate their reinforcement learning training results. Therefore, we start our study from an Early Preview Reinforcement Learning (EPRLI) algorithm built on the open-source GRPO framework, incorporating difficulty-aware intervention for math problems. Applied to a $1.5B$-parameter LLM, our method achieves 50.0\% on AIME24, 89.2\% on Math500, 77.1\% on AMC, 35.3\% on Minerva and 51.9\% on OBench—super-pass O1-Preview and is comparable to O1-mini within standard school-lab settings.
\end{abstract}

\section{Introduction}
Large language models (LLMs) like OpenAI's o-series \cite{openai2025o3mini,openai2025o4mini}, DeepSeek R1 \cite{guo2025deepseekr1}, Claude 3.7 \cite{anthropic2025claude37}, Grok-3 \cite{xai2024grok3}, and Gemini 2.5 \cite{google2025gemini25flashpreview} excel at complex reasoning tasks such as math and code generation. These capabilities are often acquired via large-scale reinforcement learning (RL), incorporating strategies like step-by-step reasoning \cite{wei2022cot}, self-reflection \cite{wang2023selfconsistency}, and backtracking \cite{ahmadian2024reinforce}. However, enhancing reasoning in small models remains difficult. To address this, we propose a preview difficulty-aware intervention-based RL algorithm that improves the math reasoning ability in small-sized large language models. Our $1.5B$ LLM, trained with an early preview of the proposed GRPO algorithm with difficulty-aware intervention, outperforms OpenAI’s O1-Preview and O1-mini \cite{openai2024reasoning, jaech2024o1} on major math reasoning benchmarks \cite{guo2025deepseekr1, christiano2017deep,everitt2021tampers, weng2024rewardhacking}. 

\section{Related Work}
\subsection{Reasoning Large Language Models}
Reinforcement learning (RL) has been widely applied to align LLMs with human preferences \cite{christiano2017deep, ouyang2022training, yuan2024preftree, azar2024theory, rafailov2023dpo}, while the open-source community has mainly relied on imitation learning \cite{yuan2024freeprocess, yue2023mammoth, guan2025rstar} to improve reasoning. Recently, the trend has shifted toward RL, with OpenAI o1 \cite{jaech2024o1} demonstrating its promise, and later works confirming the scalability of outcome-reward-based RL \cite{guo2025deepseekr1, qwen2024qwq32b, xai2024grok3}. Despite this, dense reward methods remain underexplored, as highlighted by PRIME \cite{cui2025process}, while most RL applications still use outcome reward models (ORMs) \cite{rafailov2023dpo, shao2024deepseekmath, guo2025deepseekr1}. Top-performing models—OpenAI’s o-series \cite{jaech2024o1, openai2024reasoning, openai2025o3mini, openai2025o4mini}, DeepSeek R1 \cite{guo2025deepseekr1}, Claude 3.7 \cite{anthropic2025claude37}, Grok-3 \cite{xai2024grok3}, and Gemini 2.5 \cite{google2025gemini25flashpreview}—excel in reasoning tasks. However, deep reinforcement learning are not well studied for boosting reasoning in math problems in small LLMs (0.7B–1.5B) trained on limited math data with difficulty-aware intervention.

\section{Method}

\subsection{Early Preview Group Relative Policy \cite{shao2024deepseekmath} Optimization(GRPO) with Difficulty-Aware Intervention}
We define a discrete-time finite-horizon discounted Markov decision process (MDP) by a tuple $M = (S, A, \mathcal{P}, r, \rho_0, \gamma, H)$, where $S$ is a state set, $A$ is an action set, $\mathcal{P} : S \times A \times S \to \mathbb{R}_+$ is the transition probability distribution, $\gamma \in [0, 1]$ is a discount factor, and $H$ the horizon. Our objective is to find a stochastic policy $\pi_\theta$ that maximizes the expected discounted return within the MDP, $\eta(\pi_\theta) = \mathbb{E}_\tau\left[\sum_{t=0}^{H} \gamma^t r(s_t, a_t)\right]$. We use $\tau = (s_0, a_0, \ldots)$ to denote the entire state-action trajectory, where $s_0 \sim \rho_0(s_0), a_t \sim \pi_\theta(a_t|s_t), s_{t+1} \sim \mathcal{P}(s_{t+1}|s_t, a_t)$. 

In this work, we propose a method to learn a hierarchical policy and efficiently adapt all the levels in the hierarchy to perform a new task. We study hierarchical policies composed of a higher level, or manager $\pi_{\theta_{high}}(a_{t^{high}}|s_{t^{high}})$, and a lower level, or sub-policy $\pi_{\theta_{low}}(a_{t^{low}}|s_{t^{low}})$. Both the higher level and the lower level take actions in the environment directly. The manager typically operates at a lower frequency than the sub-policies.

It is important to note that the hierarchical architecture in our preview version framework is composed of \( L \) discrete levels, where each level is indexed by \( l \in \{0, 1\} \). In this configuration, a higher-level policy corresponds to \( \text{high} = l + 1 \) and its corresponding lower-level policy is defined as \( \text{low} = l \). This structure allows for top-down coordination, in which, higher levels guide the behavior and planning strategies of the lower ones.

\begin{equation}
\label{eq:1}
\begin{aligned}
&\mathcal{J}_{\text{GRPO}^{Her}}(\theta)= \prod_{l=1}^{2}\mathbb{E}_{q^l \sim D_q^l,\, \{o_i^{l}\}_{i=1}^{G^l} \sim \pi_\theta(\cdot \mid q^{l})}\\
&\prod_{l=1}^{L}\Bigg[\frac{1}{\sum_{i^l=1}^{G^l} |o_i^l|} \sum_{i^l=1}^{G^l} \sum_{j^l=1}^{|o_i^l|} \min \Bigg(\frac{\pi_\theta(o_i^l \mid q^l)}{\pi_{\theta_{\text{old}}}(o_i^l \mid q^l)} A^{l}_{i^l,j^l}, \\
&\text{clip} \left(\frac{\pi_\theta(o_i^l \mid q^l)}{\pi_{\theta_{\text{old}}}(o_i^l \mid q^l)},\, 1-\varepsilon^{l}_{\text{low}},\, 1+\varepsilon^{l}_{\text{high}}\right) A^{l}_{i^l,j^l}\Bigg)\Bigg]
\end{aligned}
\end{equation}

In the early preview version of the proposed Group Relative Policy \cite{shao2024deepseekmath} Optimization(GRPO) with Difficulty-Aware Intervention, we adopt a simplification of the underlying Markov Decision Process (MDP). Specifically, we assume that the policies across the high and low levels share the same parameterization. This simplification leads to the expression \( \pi_{\theta_{\text{high}}}(a_{t^{\text{high}}} \mid s_{t^{\text{high}}}) = \pi_{\theta_{\text{low}}}(a_{t^{\text{low}}} \mid s_{t^{\text{low}}}) = \pi_{\theta} \), where both high- and low-level policies are treated uniformly under the shared policy \( \pi_\theta \). This unified parameterization not only reduces the model complexity but also facilitates efficient training and inference within the hierarchical structure.

\subsubsection{Reformulation}
Then, we propose the reformulation of the early preview group relative policy \cite{shao2024deepseekmath} optimization(GRPO) with difficulty-aware intervention (EPRLI) samples a group of outputs \(\{o_i^{l}\}_{i=1}^{G^{l}}\) for each question $q_{i}^{l}$ paired with the answer \(a^{l}\), $l=\{0,1\}$ and optimizes the policy via the objective represented as equation \ref{eq:1}, where $L$ is the total number of levels in the early preview group relative policy \cite{shao2024deepseekmath} optimization(GRPO) with difficulty-aware intervention (EPRLI) algorithm \ref{alg:1}. Similarly, $l$ denotes the index of the lever in the $L=2$ hierarchy, which means there are total $L=2$ hierarchy in EPRLI. 

\begin{figure*}[!t]
\centering
\begin{minipage}{0.90\textwidth}
\begin{algorithm}[H]
\caption{Early Preview EPRLI: Early Preview Group Relative Policy Optimization(GRPO) with Difficulty-Aware Intervention}
\label{alg:1}
\begin{algorithmic}[1]
\Require
    initial policy model $\pi_\theta$; reward model $\{R^l\}$; task prompts $\{\mathcal{D}^{l}\}$ with corresponding difficulty level $\{\mathcal{Q}^{l}\}$; hyperparameters $\{\varepsilon^{l}_{\text{low}}\}, \{\varepsilon^{l}_{\text{high}}\}$, $l=1,2,\ldots,L$, $\boldsymbol{Q^{l-1} > Q^{l}}$. Length Reward $\{\mathcal{K}^{l}\}$ with corresponding max length $\{\boldsymbol{Len_{\max}^{l}}\}$, $\boldsymbol{Len_{\max}^{l-1} = Len_{\max}^{l}}$.
\Ensure
    $\pi_\theta$
\For{$l = 1, \ldots, L$}    
    \For{$\text{step} = 1, \ldots, M$}
        \State Sample a batch $\mathcal{D}^{l}_b$ from $\mathcal{D}^{l}$
        \State Update the old policy model $\pi_{\theta_{\text{old}}} \gets \pi_\theta$
        \State Sample $G^l$ outputs $\{o^l_i\}_{i^l=1}^{G^l} \sim \pi_{\theta_{\text{old}}}(\cdot \mid q^l)$ for each question $q^l \in \mathcal{D}^{l}_b$
        \State Compute rewards $\{r{^l}_{i^l}\}_{i^{l}=1}^{G^l}$ for each sampled output $o^{l}_i$ by running $R^{l}$
        \State Filter out $o^{l}_i$ and add the remaining to the dynamic sampling buffer.
        \If{buffer size $n_b^{l} < N^{l}$}
            \State \textbf{continue}
        \EndIf
        \State For each $o^{l}_i$ in the buffer, compute $\hat{A}^{l}_{i^{l},t^{l}}$ for the $t^{l}$-th token of $o^{l}_i$
    \EndFor
    \For{$\text{iteration} = 1, \ldots, \mu^{l}$}
        \State Update the policy model $\pi_\theta$ by maximizing the GRPO+ objective combining with Length Reward $\boldsymbol{\mathcal{K}^{l}}$
    \EndFor
\EndFor
\end{algorithmic}
\end{algorithm}
\end{minipage}
\end{figure*}

\subsubsection{Implementation}
Then, we implement our proposed preview early preview group relative policy optimization(GRPO) with Difficulty-Aware Intervention(EPRLI) of reasoning LLMs. Particularly, the implementation is made to take the difficulty of the reasoning tasks in accordance with the hierarchy in the proposed early preview EPRLI. The details of the proposed implementations(Algorithm \ref{alg:1}) is represented as the following: 

\paragraph{Early Preview EPRLI(Algorithm \ref{alg:1}) Implementation}
In the implementation of the Early Preview Group Relative Policy Optimization(GRPO) with Difficulty-Aware Intervention framework, the hierarchy is structured with a total of two levels specifically designed to tackle mathematical reasoning problems. This two-level hierarchical design is crucial to effectively manage the complexity inherent in such tasks. More precisely, the quality values at different levels satisfy the relationships \( Q^{1} < Q^{2} \),  Additionally, the maximum allowable sequence lengths follow a strict increasing order such that \( \boldsymbol{Len_{max}^{2}} < \boldsymbol{Len_{max}^{1}} \). This setup enables the preview algorithm to handle math reasoning problems of varying lengths/difficulties while maintaining the preview hierarchical learning.

Furthermore, the hierarchical policies \( H^{1}, H^{2} \) are designed with a dominant influence order such that \( H^{1} \gg H^{2} \), meaning the top-level policy has the greater guiding power in the reasoning process, while the subsequent levels exert progressively less influence. So, optimizing this reasoning trajectory through carefully controlled hierarchical interactions.

\begin{table*}[h]
\centering
\caption{Model Performance Comparison}
\label{tab:model_comparison}
\begin{tabular}{lrrrrrr}
\toprule
\textbf{Model} & \textbf{MATH500} & \textbf{AIME24} & \textbf{AMC} & \textbf{Minerva} & \textbf{OBench} & \textbf{Avg.} \\
\midrule
\midrule
\multicolumn{7}{l}{\textbf{Close-Source}} \\
\midrule
\midrule
O1-Preview \cite{openai_o1_preview_2024} & 85.5 & 44.6 & -- & -- & -- & -- \\
O1-Mini \cite{openai2024o1} & 90.0 & 70.0 & -- & -- & -- & -- \\
O1 \cite{openai2024o1} & 90.4 & 71.5 & -- & -- & -- & -- \\
Claude 3.7 Sonnet (Standard) \cite{anthropic2025claude37sonnet} & 82.2 & 23.3 & -- & -- & -- & -- \\
\midrule
\midrule
\multicolumn{7}{l}{\textbf{Open-Source-Large}} \\
\midrule
\midrule
\textit{DeepSeek-R1} \cite{guo2025deepseekr1} & 97.3 & 79.8 & -- & -- & -- & -- \\
\textit{Qwen3-235B} \cite{qwen3-235b} & 94.6 & 85.7 & -- & -- & -- & -- \\
\textit{Llama 4 Behemoth} \cite{meta2025llama4} & 95.0 & 78.0 & -- & -- & -- & -- \\ 
\textit{Kimi-1.5} \cite{kimi2025k1.5} & 96.2 & 77.5 & -- & -- & -- & -- \\
\textit{Qwen 2.5-72B} \cite{qwen2024qwen2.5} & 83.1 & 30.0 & -- & -- & -- & -- \\
\textit{Phi4-Reasoning-14B} \cite{abdin2025phi4reasoning} & -- & 81.3 & -- & -- & -- & -- \\
\textit{Llama 4 Maverick} \cite{meta2025llama4maverick} & 18.0 & 64.0 & -- & -- & -- & -- \\
\midrule
\midrule
\multicolumn{7}{l}{\textbf{Open-Source-4B/7B}} \\
\midrule
\midrule
\textit{MIMO-7B} \cite{xiaomi2025mimo} & 95.8 & 68.2 & -- & -- & -- & -- \\
\textit{DeekSeek-Qwen-Distill-7B} \cite{guo2025deepseekr1} & 92.8 & 55.5 & -- & -- & -- & -- \\
\textit{Qwen3-4B} \cite{yang2025qwen3} & - & 73.8 & -- & -- & -- & -- \\
\midrule
\midrule
\multicolumn{7}{l}{\textbf{Open-Source-1.5B}} \\
\midrule
\midrule
\textit{DeepSeek-R1-Distill-QWEN-1.5B} \cite{guo2025deepseekr1} & 82.8 & 28.8 & 62.9 & 26.5 & 43.3 & 48.9 \\
\textit{STILL-3-1.5B-Preview} \cite{rucaibox2025still3preview} & 84.4 & 32.5 & 66.7 & 29.0 & 45.4 & 51.6 \\
\textit{DeepScaler-1.5B-Preview} \cite{deepscaler2025} & 87.8 & 43.1 & 73.6 & 30.2 & 50.0 & 57.0 \\
\textit{FastCuRL-1.5B-Preview} \cite{chen2025fastcurl} & 88.0 & 43.1 & 74.2 & 31.6 & 50.4 & 57.5 \\
\textit{Ours2-1.5B Algorithm \ref{alg:1}} & \textbf{89.2} & \textbf{50.0} & \textbf{77.1} & \textbf{35.3} & \textbf{51.9} & \textbf{60.7} \\
\bottomrule
\end{tabular}
\end{table*}

\section{Experiment}
To investigate the effectiveness of the two proposed implementations of the early preview hierarchical GRPO on the reasoning capabilities of large language models (LLMs), we conduct a series of experiments. The experiments are designed to provide a comparative analysis against the state-of-the-art reasoning-oriented LLMs of different parameters, in particular, DeepSeek-R1-Distill-Qwen-7B \cite{guo2025deepseekr1}, STILL-3-1.5B-Preview \cite{rucaibox2025still3preview}, DeepScaler-1.5B-Preview \cite{deepscaler2025}, FastCuRL-1.5B-Preview \cite{chen2025fastcurl} with $1.5$B parameters, Qwen3-4B \cite{yang2025qwen3}, DeepSeek-R1-Distill-Qwen-7B \cite{guo2025deepseekr1}, MIMO-7B \cite{xiaomi2025mimo} with middle-sized parameters, Llama 4 Maverick \cite{meta2025llama4maverick}, Phi4-Reasoning-14B \cite{abdin2025phi4reasoning}, Qwen 2.5-72B \cite{qwen2024qwen2.5}, Kimi-1.5 \cite{kimi2025k1.5}, Llama 4 Behemoth \cite{meta2025llama4}, Qwen3-235B \cite{qwen3-235b}, DeepSeek-R1 \cite{guo2025deepseekr1} with large-sized parameters, and closed-source reasoning models such as Claude 3.7 Sonnet (Standard) \cite{anthropic2025claude37sonnet}, O1, O1-Mini \cite{openai2024o1}, and O1-Preview \cite{openai2024o1preview}, enabling a thorough evaluation of the proposed early preview methods.  

\subsection{Experiment Setup} 
We choose DeepScaler-1.5B-Preview-16k \cite{deepscaler2025} as our base model, which is a $1.5B$ parameter model. We utilize the AdamW \cite{loshchilov2019decoupled} optimizer with a constant learning rate of $1 \times 10^{-6}$ for optimization. For the roll-outs, we set the temperature to $0.6$ and sample $16$ responses per prompt. In this experiment, we do not utilize a system prompt; instead, we add "Let’s think step by step and output the final answer within boxed{}." at the end of each problem.

\subsection{Benchmarks}
\paragraph{Math Reasoning Benchmark}
To better evaluate the trained model, we have selected five benchmarks to assess its performance: MATH 500 \cite{hendrycks2021iclrmmu}, AIME 2024 \cite{aimo2024a}, AMC 2023 \cite{aimo2024b}, Minerva Math \cite{Lewkowycz2022Minerva}, and OlympiadBench \cite{he2024olympiadbench}.

\subsection{Dataset and Evaluation Metric}
\paragraph{Math Reasoning Dataset}
The training dataset is consisted of $40K$ problems with two difficulty levels. Particularly, it is consisted of AIME \cite{aime24} (American Invitational Mathematics Examination) problems (1984-2023), AMC \cite{maa_amc} (American Mathematics Competition) problems (prior to 2023), Omni-MATH \cite{gao2024omnimathuniversalolympiadlevel} dataset and Still dataset \cite{rucaibox2025still3preview}. We set the maximum generation length for the models to $32768$ tokens and leverage PASS @1 as the evaluation metric. Specifically, we adopt a sampling temperature of $0.6$ and a top-p value of $1.0$ to generate k responses for each question, typically $k = 16$. Specifically, PASS @1 is then calculated as: $\text{PASS@1} = \frac{1}{k} \sum_{i=1}^{k} p_i$.

\subsection{Math Reasoning Experiments}
The proposed hierarchical reasoning model is evaluated against both open-source and closed-source state-of-the-art reasoning models, including O1-Preview \cite{openai_o1_preview_2024}, O1-Mini \cite{openai2024o1}, O1 \cite{openai2024o1}, Claude 3.7 Sonnet \cite{anthropic2025claude37sonnet}, and others. As shown in Table \ref{tab:model_comparison}, our $1.5B$ model achieves impressive performance across multiple benchmarks: $50.0$ Pass@1 on AIME24 \cite{maxwelljia_aime2024}, $89.2$ on MATH500 \cite{huggingfaceh4_math500}, $74.7$ on AMC23 \cite{amc2023}, $35.3$ on Minerva \cite{dyer2022minerva}, and $51.9$ on OlympiadBench \cite{he-etal-2024-olympiadbench}. These results demonstrate the model’s robust general reasoning ability across various mathematical and competition-level tasks.

Notably, the reinforcement learning training strategy with preview difficulty-aware
intervention enables our $1.5B$ model to outperform the current best-performing $1.5B$ reasoning model by $6.9$ points on AIME24 \cite{maxwelljia_aime2024}, $1.4$ points on MATH500 \cite{huggingfaceh4_math500}, $1.1$ on AMC23 \cite{amc2023}, $4.1$ on Minerva \cite{dyer2022minerva}, and $1.9$ on OlympiadBench \cite{he-etal-2024-olympiadbench} —averaging a $3.7$-point gain overall. Furthermore, it surpasses several larger parameter models, including O1-Preview \cite{openai_o1_preview_2024}, and is comparable with O1-2024-12-17 (Low) \cite{openai_o1_2024_12_17_low}.


\section{Discussion}
We initiate an exploration of reinforcement learning for improving the reasoning capabilities of large language models (LLMs) by introducing a preview difficulty-aware intervention strategy based on reinforcement learning, specifically tailored for mathematical problem-solving tasks. Despite being applied to a relatively small-scale math dataset, our approach demonstrates reasoning ability improvements: our 1.5B parameter model not only surpasses OpenAI’s O1-Preview \cite{openai_o1_preview_2024} but also approaches the performance level of the stronger O1-Mini model \cite{openai2024o1}. 

We plan to further develop the framework to support both small- and mid-sized models, with a longer-term goal of developing a unified reasoning agent that can excel across domains, including mathematical and coding tasks. To foster transparency and accelerate progress in this area, we commit to open-sourcing to provide the community with tools and benchmarks to advance the study of reasoning in LLMs under resource-efficient settings.

\bibliography{example_paper}
\bibliographystyle{icml2025}
\end{document}